\documentclass[letterpaper, 10 pt, conference]{ieeeconf}
\IEEEoverridecommandlockouts
\usepackage[backend=biber,style=numeric-comp,sorting=none,giveninits=true,maxbibnames=2]{biblatex}
\AtEveryCitekey{\clearlist{location}}
\AtEveryBibitem{\clearlist{location}}

\newcommand{\appropto}{\mathrel{\vcenter{
  \offinterlineskip\halign{\hfil$##$\cr
    \propto\cr\noalign{\kern2pt}\sim\cr\noalign{\kern-2pt}}}}}

\usepackage{amsmath, amssymb}
\usepackage{color}
\usepackage[svgnames]{xcolor}
\usepackage[colorlinks=false,urlcolor=black]{hyperref}
\usepackage{graphicx}

\definecolor{light-gray}{gray}{0.75}

\usepackage{algorithm,algorithmic}
\usepackage{bbold}

\newtheorem{theorem}{Theorem}
\newtheorem{lemma}[theorem]{Lemma}
\newtheorem{remark}[theorem]{Remark}

\makeatletter
\newcommand\fs@nobottomruled{\def\@fs@cfont{\bfseries}\let\@fs@capt\floatc@ruled
  \def\@fs@pre{\hrule height.8pt depth0pt \kern2pt}%
  \def\@fs@post{}
  \def\@fs@mid{\kern2pt\hrule\kern2pt}%
  \let\@fs@iftopcapt\iftrue}
\makeatother

  \usepackage{setspace}
  \usepackage{colortbl}
  \newcommand\ChangeRT[1]{\noalign{\hrule height #1}}

\usepackage[labelfont=bf,font=footnotesize]{caption}
\title{\LARGE \bf
Privacy-Preserving Map-Free Exploration for Confirming\\the Absence of a Radioactive Source
}

\author{Eric Lepowsky$^{1,2,\dagger}$ \quad David Snyder$^{1,3,\dagger}$ \quad Alexander Glaser$^{1,2}$ \quad Anirudha Majumdar$^{1,3}$ %
\thanks{
Corresponding Author: {\tt dasnyder@princeton.edu}.
$^{1}$Mechanical and Aerospace Engineering, Princeton University, NJ, USA.
$^{2}$Program on Science and Global Security (SGS), Princeton, NJ, USA.
$^{3}$Intelligent Robot Motion (IRoM) Lab, Princeton, NJ, USA.
$^{\dagger}$E. L. and D. S. contributed equally; other authors listed alphabetically.
Project GitHub: \href{https://github.com/elepowsky/verification}{github.com/elepowsky/verification}.
}%
}

\begin{document}

\maketitle
\thispagestyle{empty}
\pagestyle{empty}


\begin{abstract}

Performing an inspection task while maintaining the privacy of the inspected site is a challenging balancing act. In this work, we are motivated by the future of nuclear arms control verification, which requires both a high level of privacy and guaranteed correctness. For scenarios with limitations on sensors and stored information due to the potentially secret nature of observable features, we propose a robotic verification procedure that provides map-free exploration to perform a source verification task without requiring, nor revealing, any task-irrelevant, site-specific information. We provide theoretical guarantees on the privacy and correctness of our approach, validated by extensive simulated and hardware experiments.

\end{abstract}

\section{INTRODUCTION}
\label{sec:intro}

Autonomous robots observe and process information from their operating environments, giving rise to privacy concerns. Privacy is commonly discussed in the context of protecting personal information -- relating to a person's identity, behaviors, or health -- for instance, in assistive, social, and home robotics. Similar concerns also may arise when considering remote monitoring of mutually distrustful parties, where sensitive or compromising information must be protected \cite{tobisch}. In this work, we consider the future of nuclear arms control, which is predicated on the collection of information to verify compliance with agreed-upon limits while contending with the secrecy of the nuclear enterprise, as one such sensitive and high-stakes setting which demands privacy \cite{comley, dtra}.

One method for achieving privacy is ``forgetting,'' either by deleting previously acquired information or by compression and abstraction of the information \cite{aylett, dwork_algorithmic_2013}. For settings where the information at stake requires even more protection, such as in arms control, and inspired by the concept of ``forgetting,'' we invoke a more extreme alternative for achieving privacy: never learning or remembering in the first place. In this regard, techniques from reduced- and minimal-information decision making can help to achieve a higher level of privacy by not requiring, nor revealing, any sensitive information or any information which is not strictly task-relevant.

\begin{figure}[tp]
\centering
\includegraphics[width=\columnwidth]{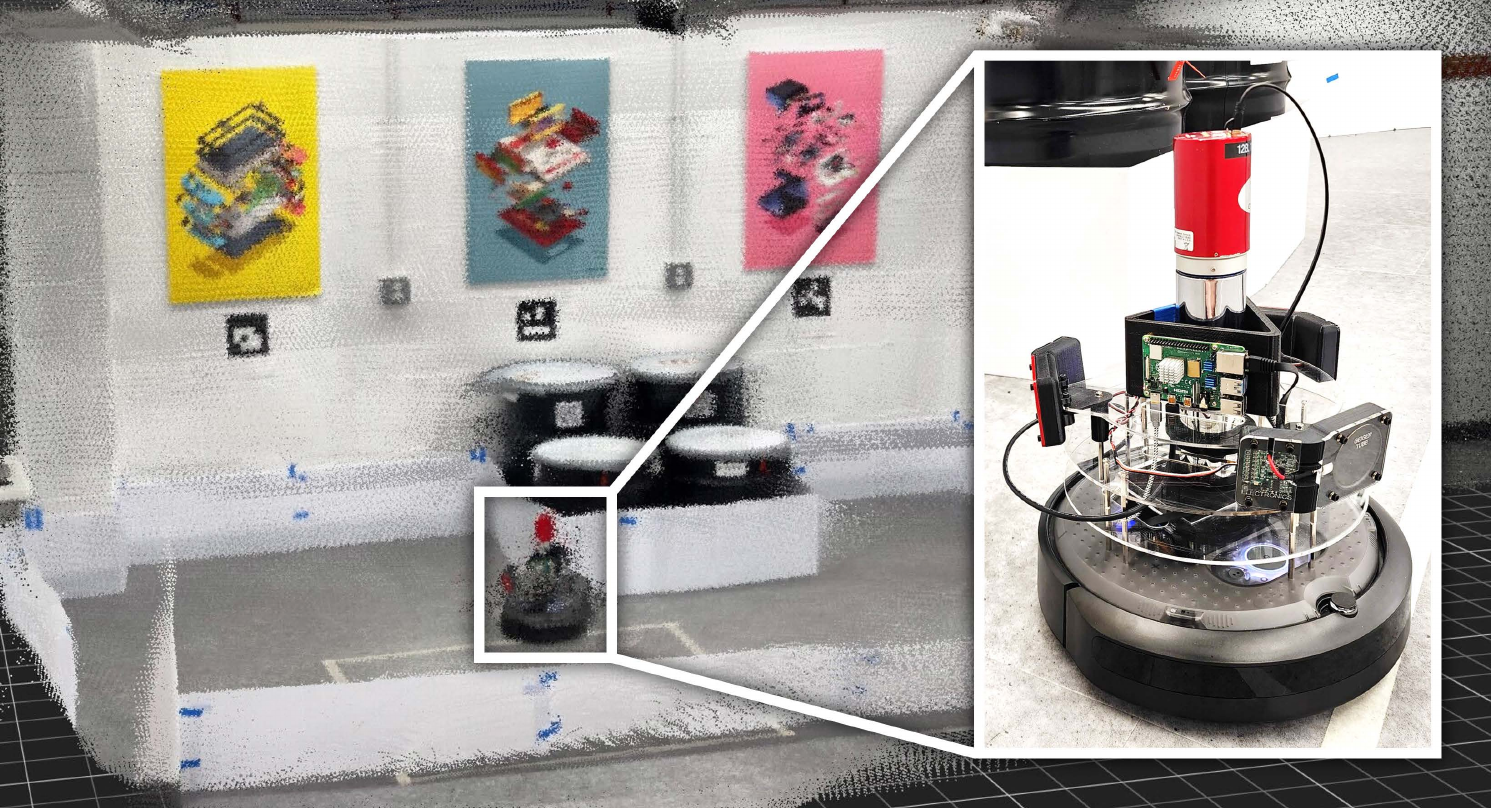}
\caption{\textbf{Robotic inspector in a representative laboratory search environment.} The environment is approximately 15~m$^2$ with steel drum obstacles and dividers to reconfigure the space. The robotic inspector, a Create 3 platform fitted with gamma-ray detectors, explores the unknown environment and confirms, with high probability, the absence or presence of a radioactive source using only non-sensitive information.}
\vspace{-4mm}
\label{fig:photo}
\end{figure}

While it is difficult to anticipate the objectives of future international agreements, they are likely to require new verification approaches which preserve aspects of onsite inspections -- traditionally essential in monitoring and verification -- while resolving concerns about intrusiveness \parencite{nas}. Onsite inspections commonly include radiation detection as a verification tool. In scenarios where no sources (e.g., nuclear warheads or fissile material) are declared, the inspector must confirm the declared absence of sources or identify an anomalous source, if present \parencite{lepowsky-nima-2021,lepowsky-sgs-2023}. Completing this ``absence confirmation'' task with high confidence remains challenging, particularly within this high-privacy paradigm.

The introduction of robotics to nuclear verification has the potential to fundamentally transform relevant inspection approaches \cite{dean, robertson, schneider, bird}, including by affording a higher level of privacy for the inspected party. Consider what a human inspector may observe during an onsite inspection: as they survey a site, they can mentally catalogue its contents, acquire measurements, or more, inevitably seeing and learning things that are sensitive yet irrelevant to the inspection task at hand. Limiting access to these observations is impossible with human-based inspections, but it can be accomplished with a carefully designed robotic inspector.

Accordingly, we address the absence confirmation task while considering any observable, site-specific, task-irrelevant information from the search environment to be sensitive (and hence limited), including imagery, dimensions, and even the site layout since mapping would reveal design characteristics of the site's contents. Many robot-compatible source detection methods, such as \parencite{mascarich, west, rahman, groves}, are in contention with such a minimal-information constraint by either using \textit{a priori} knowledge of the environment or, by consequence of the algorithmic design, revealing the site's configuration or radiation field. By considering this stringent information constraint, and demonstrating the plausibility of the inspection task in a provably-private and correct manner, we contribute to a dialogue on what may be possible in the future.

{\bf Statement of contributions.}
The primary contribution of this work is the development of a verification algorithm, which guarantees exploration while encoding only non-sensitive information, to confirm the absence of sources. We provide theoretical guarantees on privacy preservation and bounds on the false positive rate, and characterize the false negative rate in terms of parameters fundamental to the verification task. The proposed algorithm is validated in simulation and through extensive hardware experiments.

\section{RELATED WORK}
\label{sec:related}

In developing a framework for accomplishing privacy-preserving absence confirmation, we take inspiration from work on reduced- or minimal-information decision making. Techniques like dimensionality reduction \parencite{kingma_introduction_2019, vincent_stacked_2010, kingma_auto-encoding_2022, booker23} and control-theoretic methods \parencite{pacelli_robust_2022, booker_learning_2021} promote robustness via lossy, compressed representations of the sensory feedback and regularization of information usage. Analytical frameworks capturing the notion of ``available information'' quantify and formalize a relation between available and utilized information and performance \parencite{xu_theory_2020, majumdar_fundamental_2023}. While these methods are generally concerned with improving robustness and preventing overfitting, our task here is to set information usage to be absolutely minimal, then characterize the feasible system performance. From this vantage point, the literature on differential privacy is closely related in providing sufficient notions of information security \parencite{dwork_algorithmic_2013, dwork_calibrating_2006, mcsherry_mechanism_2007}; however, this does not provide guarantees on minimal algorithm performance.

To provide map-free exploration, the algorithm we propose leverages random walk processes. Of particular interest for guaranteed exploration is the cover time of random walks, or essentially how long it takes for a random walker to visit all regions of a given domain. Worst-case results for the expected cover time on undirected graphs have been given in \parencite{aldous,aleliunas}. General properties for finite graphs of various structures have been shown in \parencite{aldous_strong_1987, mihail_conductance_1989, sinclair_approximate_1988, ball_mean_1997}, with classical Perron-Frobenius theory summarized in \parencite{seneta_non-negative_1981, serre_matrices_2010}. There has also been interest in developing approximations for expected cover times, as in \parencite{dong, chupeau_cover_2015, regnier_universal_2023}. For sampling from sparse fields, L\'evy flights, which exhibit heavy-tailed distributions over step size, are one means of efficient, multi-scale, bio-inspired exploration \parencite{viswanathan, reynolds, pang}. While this particular type of random walk could expedite exploration, high-probability coverage properties would require additional analytical scaffolding, complicating the use of such multi-scale strategies.

\section{PROBLEM FORMULATION}
\label{sec:problem}

The problem is twofold, requiring definitions for both the search environment and the source verification task. Although the goal is to confirm the absence of sources, this framework requires that a source must be detected, if present.

\subsection{Defining the Environment}

Assume that a site is declared to contain no sources. The inspection task is to verify the declaration by confirming the absence of sources (or their presence in a non-compliance situation) by traversing the free space and measuring the observed scalar field. Complicating this is our information constraint: the capacity of the robotic inspector to retain information must be kept to a minimum. Ideal verification methods must provide both \emph{calibrated correctness} (the ability to choose the probability of returning the correct inspection result) and \emph{provable privacy} (minimizing the robot's capacity to ``leak'' information). Notably, these goals are generally in opposition; correctness would benefit from more information to better characterize the space, while privacy would require less information be available to the robot.

For the search environment $E(\mathcal{I}, s, M)$, the robotic inspector $\mathcal{I}$ is tasked with determining the absence or presence of a source of emission strength $s \geq 0$ in the map $M$, where $s = 0$ corresponds to ``no source''. The robot $\mathcal{I}$ has a fundamental length $r_I > 0$ (e.g., its diameter). The map $M(l_x, l_y, B)$ is physically bounded by positive length constants $l_x, l_y$ with an unknown occupancy function defining the free space. $l_x, l_y$ are assumed to be non-sensitive if they may be determined without access to the search environment, for instance through open-source information or outside observation. For radiation detection, the environment also has a Poisson-distributed background of mean $B \geq 0$.

\subsection{Map Compression}

For this work, we must restrict ourselves to the class of maps with a single, traversable (i.e., contiguous) region of free space. We regularize the set of valid maps by discretizing the problem into a directed graph representation, where each node is a region of space. A full description of the discretization procedure is provided in App.~\ref{app:discretize}. Specifically, the inequalities of Eq.~\ref{Eqn:CompressionInequalities} must hold for the discretization length $\epsilon_M$ of map $M$, where $r_D$ is the detector range, or the distance from which a source is readily detectable above background. We conservatively take $r_D$ to be the distance at which the signal-to-background ratio reaches unity.

\begin{equation}
\label{Eqn:CompressionInequalities}
    r_I \leq \epsilon_M \leq \frac{r_D}{\sqrt{2}}
\end{equation}

The left-hand inequality ensures traversability after discretization, while the right-hand inequality ensures that if the robot enters a particular bin, it can detect a source from anywhere in that bin. This latter statement implicitly assumes that sources are detectable, which is dependent on various factors, including any intervening material between the source and the detector; this problem has been explored in the context of radiation detection for treaty verification, for example in \parencite{lepowsky-nima-2021}, and will be momentarily neglected, such that excessively shielded sources are considered to be incompatible with the present formulation. Therefore, each time the discretized space is covered, there must be at least one potential anomalous measurement if a source is present.

Henceforth, we will refer to a \emph{compressed map} as $M(l_x, l_y, B, \epsilon_M)$ and define a class of compressed maps as $\mathbb{M}(l_x, l_y, B, \epsilon_{\underline{M}}) = \{M(l_x, l_y, B, \epsilon_M) : \epsilon_{M} \geq \epsilon_{\underline{M}}\}$. The property $\epsilon' \geq \epsilon \implies \mathbb{M}(l_x, l_y, B, \epsilon') \subseteq \mathbb{M}(l_x, l_y, B, \epsilon)$ follows directly. Incorporating Eq.~\ref{Eqn:CompressionInequalities} with this definition, the valid set of maps for a given inspector is the set difference given by $\mathbb{M}_\mathcal{I} = \{\mathbb{M}(l_x, l_y, B, r_I) \setminus \mathbb{M}(l_x, l_y, B, \frac{r_D}{\sqrt{2}})\}$. 

Critically, although the inspector acts in $M \in \mathbb{M}_\mathcal{I}$, it does not see nor construct a representation of the underlying map. Furthermore, to maintain the privacy of the site, the inspector does not collect or store information (e.g., a state history) which would be sufficient to deduce the map, nor does it store information (e.g., scalar measurements) which would be sufficient to characterize the radiation field. We emphasize that all map-dependent results (coverage times and fractions) are from an omniscient view unavailable to the inspector.

\subsection{Source Detection with Limited Information}

The verification task has two distinct failure modes: a false negative occurs if the robotic inspector incorrectly returns ``absence confirmed,'' and a false positive occurs if the robotic inspector incorrectly returns ``anomaly detected.'' To minimize the false negative rate (FNR), the robot needs to guarantee exploration of the space, such that a source would be detected, if present. Additionally, each indicator of source-presence must individually have a guaranteed false positive rate (FPR). This is similar to a standard suite of problems in robotics, including out-of-distribution detection, anomaly detection, and failure prediction \parencite{sharma_sketching_2021, sinha_system-level_2022, farid_failure_2022, farid23, luo22}. What distinguishes our setting is the fundamental constraint that the stored information $\mathcal{G}_t$ be exclusively non-sensitive, that it not allow for reconstruction of the underlying map, and that this property holds uniformly across all $t \in \mathbb{N}$.

Given $r_I$ and $r_D$, it is assumed that the map $M$ is drawn from the class of valid maps $\mathbb{M}_\mathcal{I}$. This assumption is not too onerous given that human inspectors also have non-zero extent, and therefore would struggle in an overly obstacle-dense map. Granted, we cannot rival humans in being able to flag certain ``adversarial'' maps -- though they would use (sensitive) sensory information to do so. Second, the measurement model $h(x_t,y_t;E)$ must be predictable, such that anomalous detections may be differentiated from the background. Critically, the actual measurements $h_t$ are sensitive and cannot be stored directly. The robot's position also cannot be known during operation, as this would reveal information about the map. The verification algorithm $\mathcal{A}$ must take physical actions and make decisions $d_t$ (``absence confirmed,'' ``anomaly detected,'' or ``continue'') that rely only on non-sensitive accumulated information $\mathcal{G}_t$.

For radiation detection, $h \sim \mathcal{P}(B + g(s, x, y))$ is Poisson-distributed. The non-negative function $g$ is $0$ if $s=0$ or the robot position $(x, y)$ does not have line-of-sight to the source; this assumes that obstacles are completely attenuating, which is a simplification of the absorption and scattering which occurs in real-life. Otherwise, $g \appropto \frac{s}{r^2}$, for $r$ as the Euclidean distance from the inspector to the source; the inverse-square law is merely an approximation, which we replace with an experiment-based model in our presented demonstration.

\section{METHODOLOGY}
\label{sec:methodology}

The algorithm we propose takes inspiration from randomized, sampling-based motion planners \parencite{lavalle, kuffner} and out-of-distribution detection \parencite{basseville}. Our random walk policy encodes the scalar measurements as physical actions, guaranteeing exploration of the search environment while simultaneously accumulating $\mathcal{G}_t$, which is a non-sensitive proxy of the measurement history; the history itself is \textit{never} stored.

Consider a robot that can translate forward and rotate in place, detect imminent collisions, and accurately acquire scalar (radiation) measurements; such a system can run Alg.~\ref{alg:random}. When a measurement is consistent with source-absence, the robot moves according to a ``reference'' random walk (maximum step size $c_U$) that explores the space; otherwise, if consistent with source-presence, it moves according to an ``out-of-distribution'' random walk (maximum step size $c_L$). Since the actions depend only on the measured scalar, the resulting distribution over actions (step sizes) for any source-free map is theoretically identical; we refer to this source-free action distribution as the reference, $V_r$. Detection of a shift in the realized action distribution, denoted $V_e$, is accomplished by Kolmogorov-Smirnov (KS) testing \parencite{kolmogorov}. We set the confidence parameter $p^*$ based on the KS test P-value to determine if the distributions are more likely distinct.

\begin{algorithm}[htbp]
\caption{Random walk absence confirmation.}
\begin{algorithmic}
\label{alg:random}
\STATE \textbf{Input}: Estimated background $B$, Outer dimensions $l_x, l_y$, Confidence parameter $p^*$, Run time $T$, Test count $n$, Threshold level $z$, Step size constants $0 \leq c_L < c_U$, Reference distribution $V_r$ \\
\STATE \textbf{Output}: Inspection result \\
\STATE \textbf{Initialize} P-value $\underline{p} =1.0$, Time step $t = 1$, Realized action distribution $V_e = \{\emptyset\}$, Starting pose $x_0, y_0, \theta_0$ \\
\WHILE {$t \leq T$}
    \STATE $N_t \sim h(x_t, y_t; E)$ \hfill \algorithmiccomment{Field measurement} \\
    \STATE $c \gets c_L + (c_U-c_L)\mathbb{1}[N_t \leq B + z\sqrt{B}]$ \hfill \algorithmiccomment{Set max step} \\
    \STATE $ds, \:d\theta \sim \mathcal{U}[0,c], \:\mathcal{U}[0,2\pi]$ \hfill \COMMENT{Step length, rotation} \\
    \STATE Rotate by $d\theta$ rad. and move forward $ds$ distance \\
    \STATE Append $ds$ to memory $V_e$
    \IF{$t \equiv 0$ (mod $T/n$)}
        \STATE $\underline{p} = \min\{\underline{p}, \mathbf{KS}(V_e, V_r)\}$ \hfill \algorithmiccomment{Perform KS test} \\
    \ENDIF
    \IF{$\underline{p} \leq p^*/n$}
        \RETURN $1$ \hfill \COMMENT{Result: Anomaly detected} \\
    \ENDIF
\ENDWHILE
\RETURN $0$ \hfill \COMMENT{Result: Absence confirmed} \\
\end{algorithmic}
\end{algorithm}

The robot takes smaller steps when anomalously high counts are detected (if the observed counts exceed $z$ standard deviations above the expected background, as is common practice in radiation detection). We set the reduced step size to $c_L = \frac{c_U}{10}$; for $c_L \lesssim \epsilon_{\underline{M}}$, the inspector is likely to stay in the vicinity of an anomalous source so that evidence of source-presence is self-reinforcing. The step size ($ds$) recorded is the randomly-selected distance between measurement points; when the robot encounters an obstacle, it randomly redirects then travels the remaining distance. Alg.~\ref{alg:random} terminates once the maximum number of steps is reached ($d_t = 0$, ``Absence confirmed'') or if the KS test P-value reaches the threshold of $p^*/n$ ($d_t = 1$, ``Anomaly detected'').

Note that Alg.~\ref{alg:random} requires an estimate of the background. For this proof-of-concept, we assume that the background has been previously characterized (for example, when the environment was first initialized). A more robust approach could consider pseudo-online learning of the background, which we defer to future work. A possible complication would be a non-uniform background field, for which more thorough characterization would violate the information constraint. However, as a workaround, the estimated background, $B$, can be re-interpreted as $B_{max}$ which captures any fluctuations, such that ``absence confirmed'' means that nothing over the maximally acceptable background is present.

\section{PRIVACY AND CORRECTNESS}
\label{sec:validation}

The proposed methodology is now characterized theoretically to confirm that it satisfies the two required properties: provable privacy and calibrated correctness.

\subsection{Information Privacy}

A satisfactory verification approach must be private, meaning that it does not ``leak'' any sensitive information, as formalized by Theorem~\ref{thm:privacy}. A full proof of Theorem~\ref{thm:privacy} is provided in App.~\ref{app:mutual}.

\begin{theorem}[Information Privacy of Compliant Hosts]
\label{thm:privacy}
Consider the class of compliant (source-free) maps, denoted by $\mathbb{M}^-(l_x, l_y, B, \epsilon_{\underline{M}})$. Alg.~\ref{alg:random} is considered to be private, for all time, $t$, with respect to any map, $M \in \mathbb{M}^-$, in that the mutual information ($\mathcal{MI}$) between any stored data point (namely, the step size between measurements) and the particular compliant (source-free) map is zero \parencite{mutual_info}. This mutual information is mathematically expressed by Eq.~\ref{eqn:MutualInfo}, for each data point $\{\mathcal{G}_t \setminus \mathcal{G}_{t-1}\}$ (equivalent to $ds_t$) and class of maps $\mathbb{M}^-$. In other words, at no time in its operation can Alg.~\ref{alg:random} distinguish between any pair of compliant maps.
\begin{equation}
\label{eqn:MutualInfo}
\mathcal{MI}(\{\mathcal{G}_t \setminus \mathcal{G}_{t-1}\}, \mathbb{M}^-) = \mathcal{MI}(ds_t, \mathbb{M}^-) = 0 \text{ } \forall t \geq 1
\smallskip
\end{equation}
\end{theorem}

\begin{figure}[htbp]
    \includegraphics[width=0.93\columnwidth]{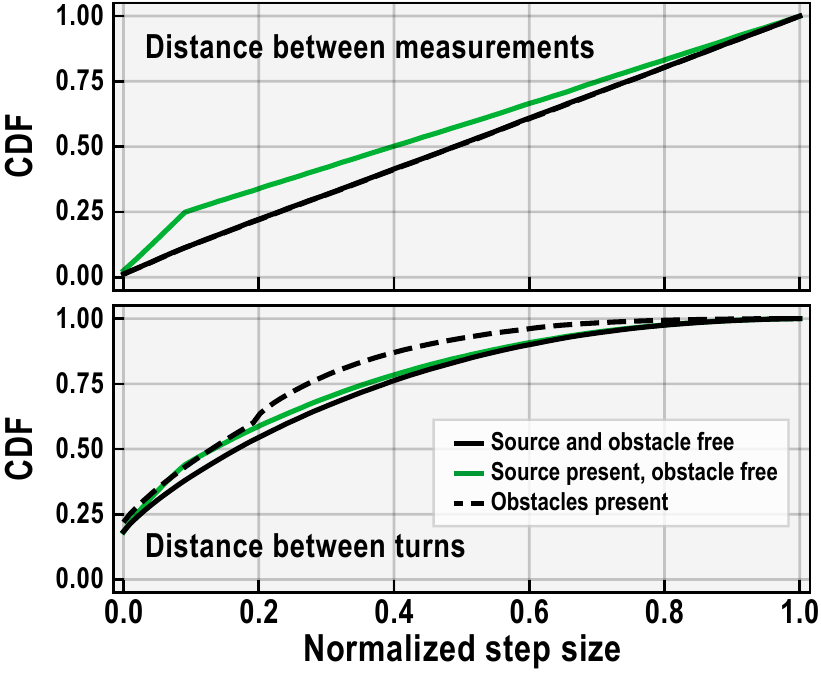}
    \caption{\textbf{Step size distributions for information storage scheme privacy and counterexample.} Cumulative density functions over step size for our algorithm (distance between measurements) and a ``leaky'' alternative (distance between turns). Our algorithm (above) is only dependent on the presence/absence of a source, whereas the seemingly similar information storage scheme (below) leaks information which can differentiate between environments of differing occupancy. Note that the solid and dashed black lines in the upper plot are overlapped; this particular curve is equivalent to the reference distribution, $V_r$, which is independent of the environment.}
    \label{fig:cdf}
    \vspace{-3mm}
\end{figure}

To further demonstrate this result, Figure~\ref{fig:cdf} shows (in simulation) that Alg.~\ref{alg:random}, which stores only the step size between measurements, yields a result that depends \emph{only} on the absence or presence of a source. The equivalency between the realized action distribution for empty and obstacle-filled environments concretely demonstrates that there is zero mutual information, per Theorem~\ref{thm:privacy}. Conversely, as a counter-example, a seemingly similar information storage scheme which stores the step size between turns ``leaks'' enough information to differentiate between environments.

\subsection{Inspection Correctness}

Correctness requires guaranteeing a low false negative rate (FNR) without compromising the false positive rate (FPR), and vice versa. We begin with an immediate characterization of the false-positive calibration by the setting of $p^*$ and $n$.

\begin{remark} [Calibrated False Positive Rate] The FPR of Alg.~\ref{alg:random}, equivalently the probability of incorrectly detecting an out-of-distribution anomaly, is less than or equal to $p^*$. This follows from a union bound applied to the outcomes of $n$ pre-specified Kolmogorov-Smirnov (KS) tests, each individually performed at a significance of $p^*/n$.
\end{remark}

To address the FNR, full coverage of the environment with sufficiently high resolution is necessary to eliminate the possibility of source-presence. Unfortunately, standard coverage algorithms typically rely on detailed knowledge of the environment \parencite{cao,khanam,sadat}. Even planners which don't require a map \emph{a priori} typically maintain a state history, forming a representation of the space that is incompatible with the minimal-information constraint. Although less time-efficient, we use random walk processes to provide the necessary exploration without requiring any environmental information.

In essence, our absence confirmation algorithm is a random walk in continuous space, reduced to a discrete graph for tractability. To calibrate the FNR, we require a bound on coverage time ($T$) that guarantees, with high probability, full coverage of the environment (i.e., every discretized bin has been visited). Formally, we desire a function $\mathcal{T}(N) = \max_{M} T_M \: \forall M \in \mathbb{M}(l_x, l_y, B, \epsilon_{\underline{M}})$ which provides an upper bound on coverage time for a given class of maps. The following lemma characterizes the tail behavior of the distribution governing this quantity. A full proof of Lemma~\ref{lemma:exponential} is provided in App.~\ref{app:exponential}.

\begin{lemma}[Passage Times in Exponential Family \parencite{mihail_conductance_1989}]
\label{lemma:exponential}
Consider any compressed map with $N$ nodes and graph diameter $D_G$, radius $r_G$. The distribution of first passage times to node $i$ from any other node $j \neq i$ is a member of the exponential family; the first passage time from node $j^* \neq i$ to node $i$ is distributed geometrically in non-dimensionalized time $\tau = \frac{t}{r_G(i)}$, where $r_G(i) \leq D_G \leq N$. This ensures that relatively tight high-probability bounds on coverage can be obtained (as exemplified by the analysis provided in App.~\ref{app:convolution}); it also reflects how the particular structure overcomes several worst-case coverage time results.
\end{lemma}

Achieving a calibrated FNR also requires quantifying the conditional probability of source detection \emph{given} full graph coverage. We assume that potential sources are sufficiently strong (or the detector is sufficiently sensitive within $r_D$) such that the conditional probability is essentially equal to one, which is consistent with the simulations and hardware experiments. Although simplified here, the notion of detectability is a multi-faceted yet analytical problem \parencite{knoll}; generally, notwithstanding excessive shielding, higher efficiency detectors and longer sampling periods improve detectability.

Backed by these theoretical underpinnings, we now assess the coverage time empirically. A diverse set of 10 simulated environments were utilized (see Section~\ref{sec:simulation}), several of which reflect known worst-case configurations for undirected graphs, such as a barbell graph. We simulate 50 independent trials for each $10\times10$~m environment and each of 5 different maximum step sizes, $c_U=$ (2, 4, 6, 8, 10~m). The coverage versus time for a range of discretization sizes (25, 100, or 400 bins of corresponding side length 2, 1, or 0.5~m) is summarized in Table~\ref{tab:coverage}. We can use these results to approximate the upper bound on coverage time, $\mathcal{T}(N)$.

\begin{table}[htbp]
\renewcommand{\arraystretch}{1.3}
\footnotesize
\begin{center}
\begin{tabular}{>{\centering\arraybackslash}p{1.4cm} !{\vrule width 1pt} >{\centering\arraybackslash}p{0.9cm} | >{\centering\arraybackslash}p{0.9cm} | >{\centering\arraybackslash}p{0.9cm} | >{\centering\arraybackslash}p{0.9cm} | >{\centering\arraybackslash}p{0.9cm}}
& 2~m & 4~m & 6~m & 8~m & 10~m
\\
\ChangeRT{1pt}
5$\times$5 bins & 810 (3529) & 305 (1861) & 200 (1721) & 159 (818) & 145 (699)
\\ 
\hline
10$\times$10 bins & 1481 (5285) & 741 (2614) & 611 (1932) & 547 (1444) & 547 (1369)
\\
\hline
20$\times$20 bins & 3422 (12161) & 2502 (6710) & 2300 (6369) & 2256 (6295) & 2232 (5022)
\\
\end{tabular}
\end{center}
\vspace{-1mm}
\caption{\textbf{Empirical coverage time (step number) for varied maximum step size and discretization.} The mean number of steps, averaged over all 10 environments and all 50 trials, is reported; the maximum over all rooms and trials is reported in parentheses to demonstrate the worst case observed.}
\label{tab:coverage}
\end{table}

\begin{figure}[htbp]
    \includegraphics[width=0.93\columnwidth]{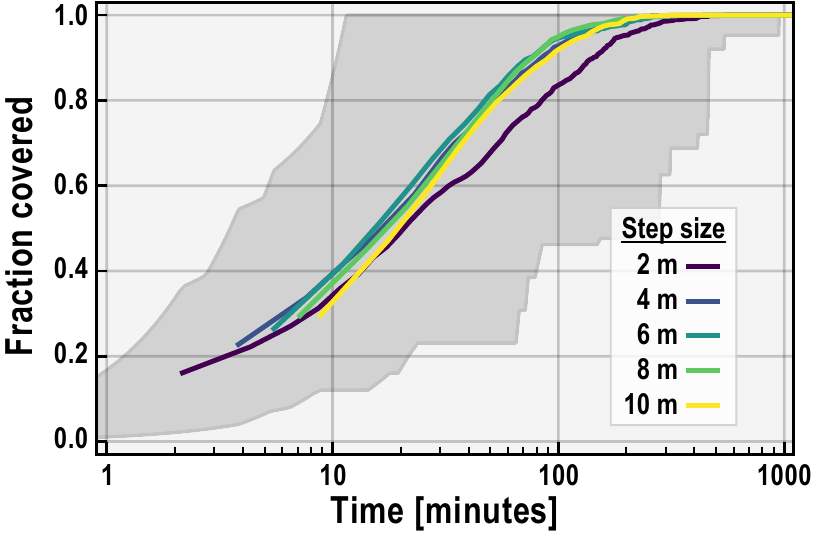}
    \caption{\textbf{Empirical coverage versus time.} Evaluated for a range of maximum step sizes for the 5$\times$5 binning. For each step size, the average over all 10 environments and all 50 trials is shown; the curves start after 10 initial time steps. The shaded region represents the full range of possible values, evaluated over all step sizes. To convert from step number to real-world time, we assumed 3-second measurements, travel speed of 10~cm/s, and neglect the time spent avoiding obstacles.}
    \label{fig:coverage}
    \vspace{-2mm}
\end{figure}

The tabulated results are intuitive: neglecting travel time, larger maximum step sizes and larger bins (equivalently, fewer bins) result in lower coverage time. To further elucidate the results from Table~\ref{tab:coverage}, Figure~\ref{fig:coverage} visualizes the empirical coverage over time (converting from number of steps to real-world time) for the 5$\times$5 binning; this particular size was chosen since the 25-bin compressed maps are most closely aligned with the bin number of the hardware experiments.

\section{EXPERIMENTS}
\label{sec:experiments}

We experimentally demonstrate the correctness of our algorithm, i.e., the ability to correctly identify the absence or presence of a source, both in diverse simulated environments and on hardware in various laboratory settings, collectively spanning a wide range of scales and configurations.

\subsection{Simulation in PyBullet}
\label{sec:simulation}

\begin{figure}[hptb]
    \centering
    \includegraphics[width=\columnwidth]{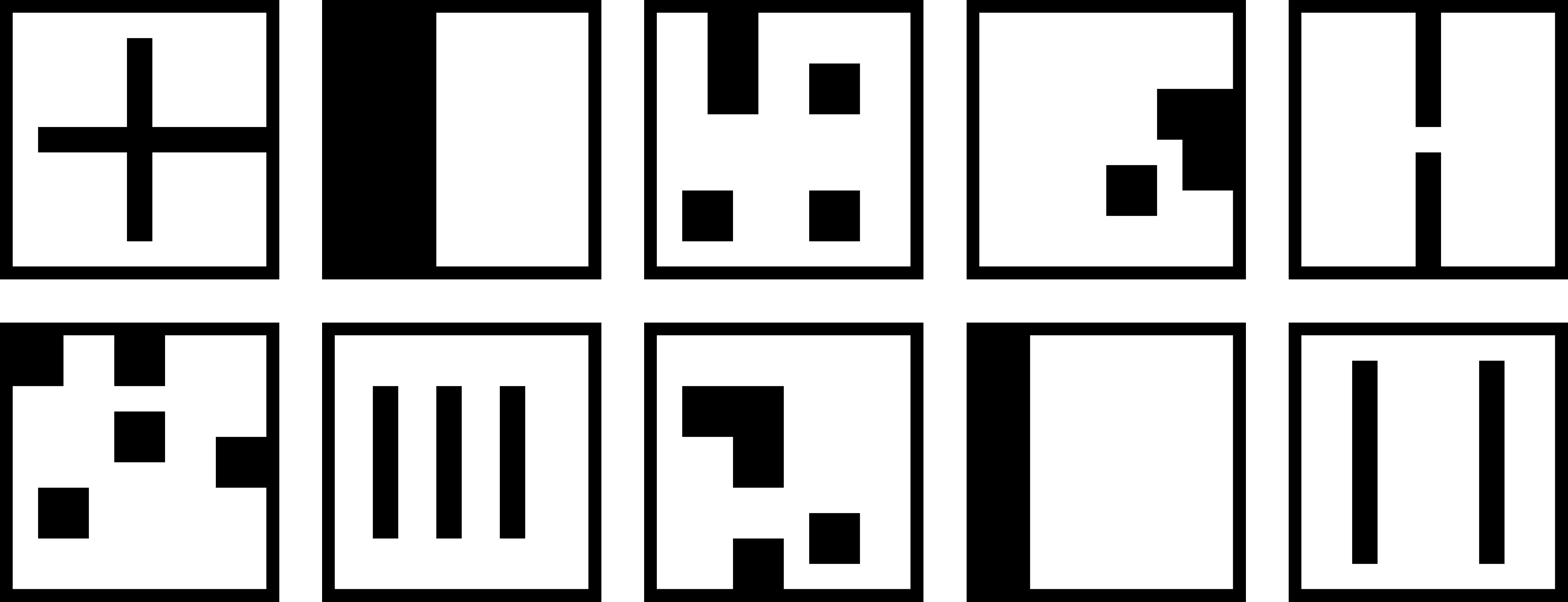}
    \caption{\textbf{Selection of simulated environments with varying complexity.} All rooms are designed with the same 10$\times$10~m outer dimensions with different occupancy fractions and obstacle shapes.}
    \label{fig:environments}
\end{figure}

Our simulation environment uses PyBullet \parencite{pybullet,pybulletgym}, based on the environment setup from \parencite{wheeled}. A variety of environments were constructed (30 in total, in addition to the 10 unique environments used for Table~\ref{tab:coverage}), each with different occupancy functions, including an assortment of maps with open space ranging from 100~m$^2$ (an entirely empty room) down to 20~m$^2$ (obstacle-dense maps). Representative environments are shown in Figure~\ref{fig:environments}. For each map, 100 independent trials were conducted: 10 with and 10 without a source present, each for 5 different maximum step sizes. For each trial, the robot and source (if present) were initialized in random positions. Ray-tracing provided a realistic, albeit simplified, measurement model, where obstacles were assumed to be fully attenuating and the spatial dependence for non-attenuated counts was experimentally-based.

\begin{figure}[htbp]
    \centering
    \includegraphics[width=\columnwidth]{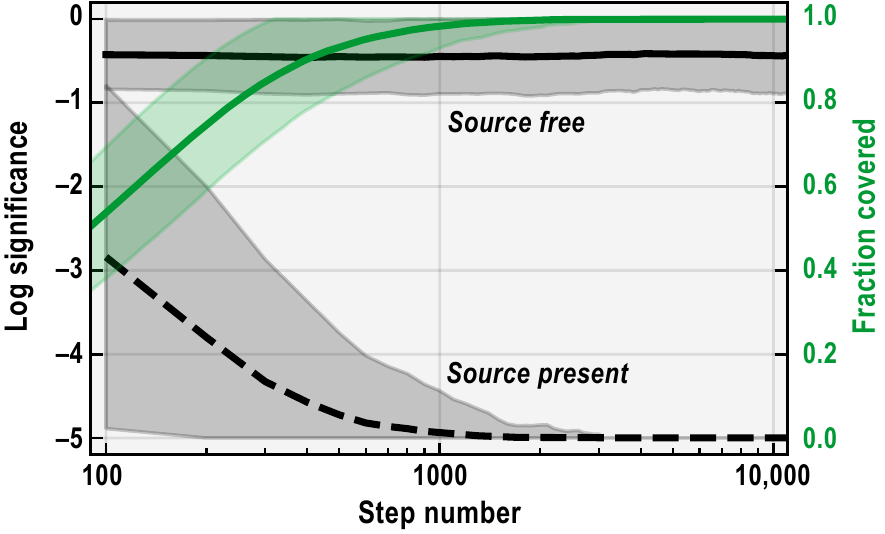}
    \caption{\textbf{Evolution of KS test significance and spatial coverage for simulated trials.} The results for source-absence (solid) and source-presence (dashed) are averaged over all trials and simulated rooms. Coverage is shown for the source-absence case; coverage for the source-presence case is omitted since the algorithm terminates once a source is confirmed. The log-significance, represented in $\log_{10}$-space, is floored at the KS test trigger threshold, which was $-5$ for the simulated trials. For each curve, the shaded region represents one standard deviation of the full range of values.}
    \label{fig:demo-simulated}
    \vspace{-1mm}
\end{figure}

The evolution of the KS test P-value for Alg.~\ref{alg:random}, averaged over all similar trials across all environments (i.e., 1,500 trials each for source absence and presence), is shown in Figure~\ref{fig:demo-simulated}. We apply the KS test after every 100 measurements, using $p^* = 0.005$ and $n = 500$, assuming a conservative upper coverage time of 50,000~steps. This yields an overall confidence of 99.5\%. For omniscient reference, the corresponding average coverage is included; this information is \emph{not} acquired or inferable given the data storage of Alg.~\ref{alg:random}. 

Inspecting Figure~\ref{fig:demo-simulated}, we see that full coverage of the environment was always achieved in the source-free case, as evidenced by the converged standard deviation, and there was never a statistically significant difference between the reference and realized action distributions (no false positives). For the source-present case, we likewise see that all trials correctly converged to the KS test trigger threshold of $\log_{10}(\frac{p^*}{n}) = -5$ (no false negatives). Coverage is not shown for the source-presence case since full coverage is not guaranteed, nor is it required, when a source is present.

\subsection{Hardware Demonstration}

We built a simple gamma ray-detecting robot (shown in Figure~\ref{fig:photo}) using the iRobot Create 3 platform, though we note that the proposed methodology is sensing agnostic and applicable to many possible scalar fields and sources. Collision avoidance is accomplished using the onboard infrared sensors. The radiation detection unit uses a 2-inch Mirion/Canberra NaI scintillator (Model 802) connected to an Osprey Digital MCA Tube Base. A Raspberry Pi 4 Model B reads the data over Ethernet and relays the counts to the controller via Bluetooth. For rudimentary filtering to reduce low energy noise, channels $\ge 400$ of the 2048-channel spectrum are summed to yield gross counts. Based on experimental calibration (i.e., counts versus distance from a source), we take the detector range to be $r_D = 1~m$.

\begin{figure}[htbp]
    \centering
    \includegraphics[width=\columnwidth]{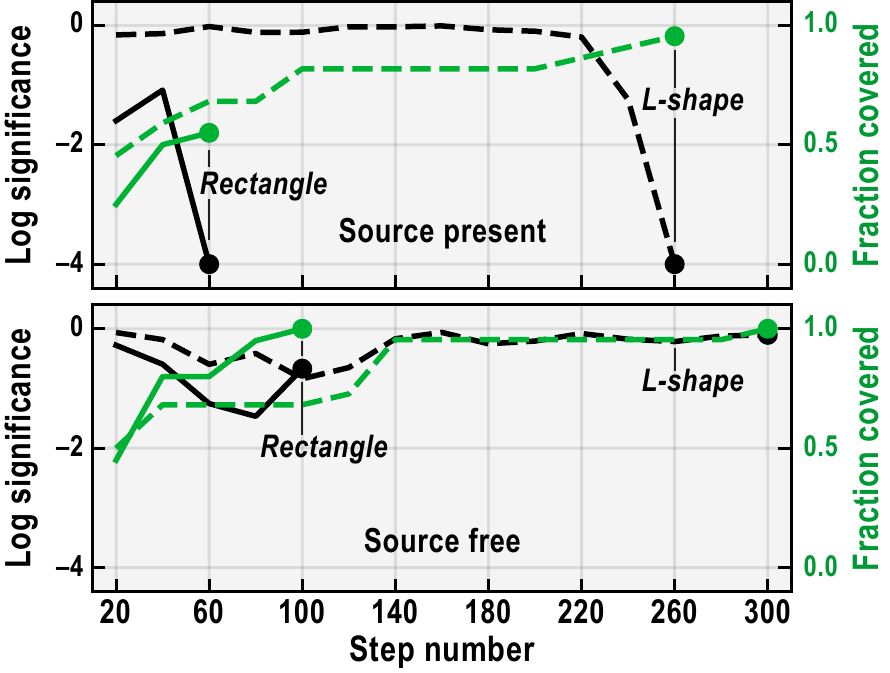}
    \caption{\textbf{Evolution of KS test significance and spatial coverage for full-scale laboratory trials.} For the source-presence cases (above), the time step when the algorithm terminates is indicated (i.e., once the KS test log-significance, represented in $\log_{10}$-space, reaches $-4$). For the source-absence cases (below), the time step when coverage is first reached is indicated; note that the algorithm would \textit{not} actually terminate in this case, since coverage is not known to the inspector.}
    \label{fig:demo-hardware}
    \vspace{-3mm}
\end{figure}

Various laboratory environments were configured. For source-present environments, a set of gamma-ray check sources were used (Cs-137, Ba-133, and Co-60, among other isotopes) totaling to around 9~$\mu$Ci of activity. The hardware analog to Figure~\ref{fig:demo-simulated} is shown in Figure~\ref{fig:demo-hardware} for trials conducted in two full-scale environments (20~m$^2$ rectangle and 22~m$^2$ L-shape). For these trials, the NaI detector was utilized with 1~m$^2$ bins for tracking coverage. We apply the KS test after every 20 measurements, using $p^* = 0.005$ and $n = 50$, assuming an upper coverage time of 1000~steps. Accordingly, for the hardware trials, the KS test trigger threshold was $\log_{10}(\frac{p^*}{n}) = -4$. As with the simulated trials, these parameters yield an overall confidence of 99.5\%.

For both environments, the KS test correctly reached the set threshold in the source-present case and full coverage was achieved without reaching the KS test threshold in the source-free case. Coverage was tracked in real-time to provide an omniscient view (unknown to the robot) of the experiment. The trials were artificially stopped once coverage was reached, but according to Alg.~\ref{alg:random}, the robot would have continued its inspection until the maximum time was reached. In addition to these large-scale environments, we conducted trials in four smaller rooms of varying complexity, summarized in Table~\ref{tab:demo-hardware}. For each environment, 10 trials were conducted: 5 with the gamma sources and 5 without the sources. For all trials in all environments, Alg.~\ref{alg:random} yielded the correct result: when no source was present, the robot covered the environment without reaching the KS test threshold; when a source was present, the robot more expeditiously reached the threshold, indicative of an anomalous source.

\begin{table}[ht]
\vspace{1mm}
\renewcommand{\arraystretch}{1.3}
\footnotesize
\begin{center}
\begin{tabular}{>{\centering\arraybackslash}p{1.25cm} !{\vrule width 1pt} >{\centering\arraybackslash}p{1.25cm} | >{\centering\arraybackslash}p{1.25cm} | >{\centering\arraybackslash}p{1.25cm} | >{\centering\arraybackslash}p{1.25cm}}
& \includegraphics[width=0.9cm]{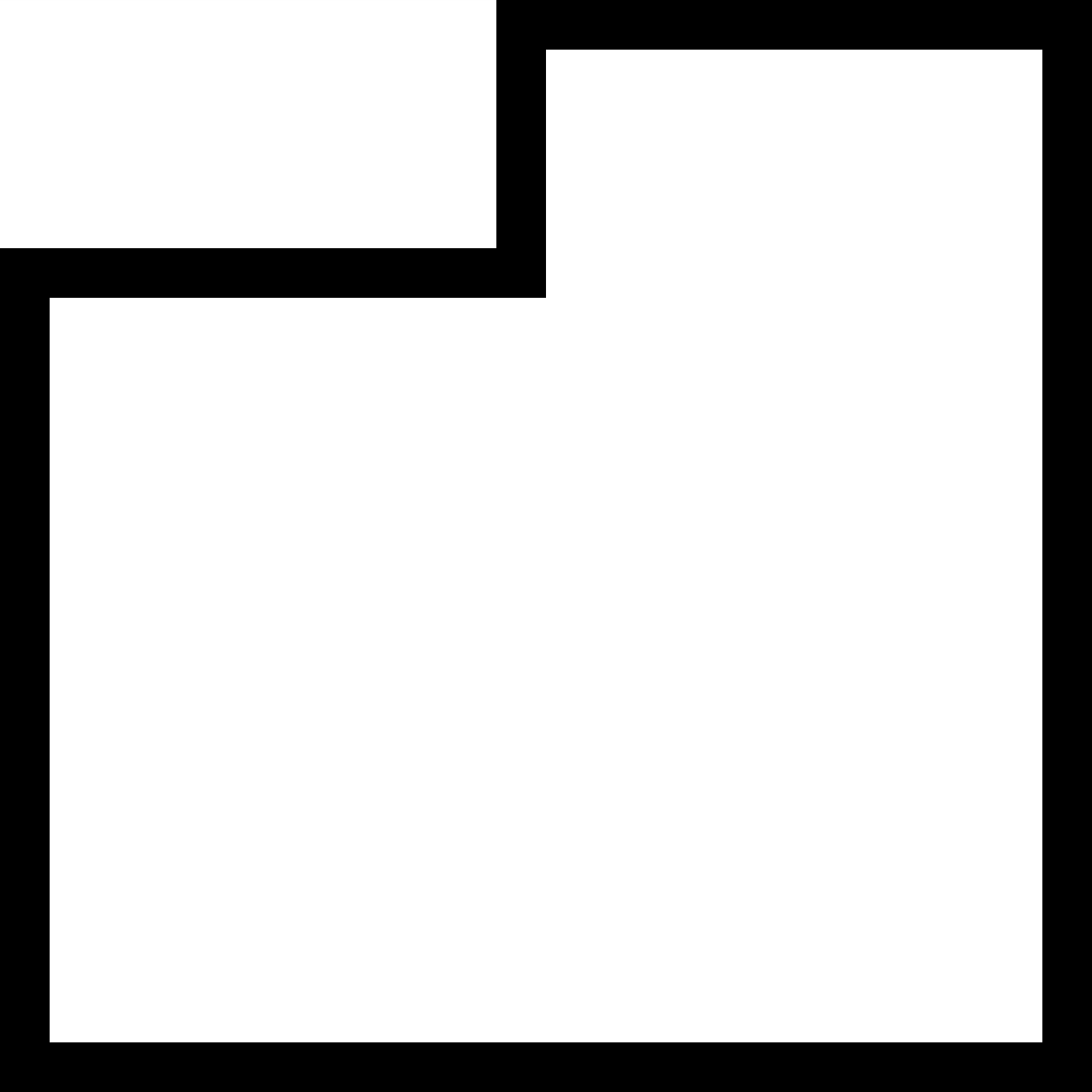} 14~m$^2$ & \includegraphics[width=0.9cm]{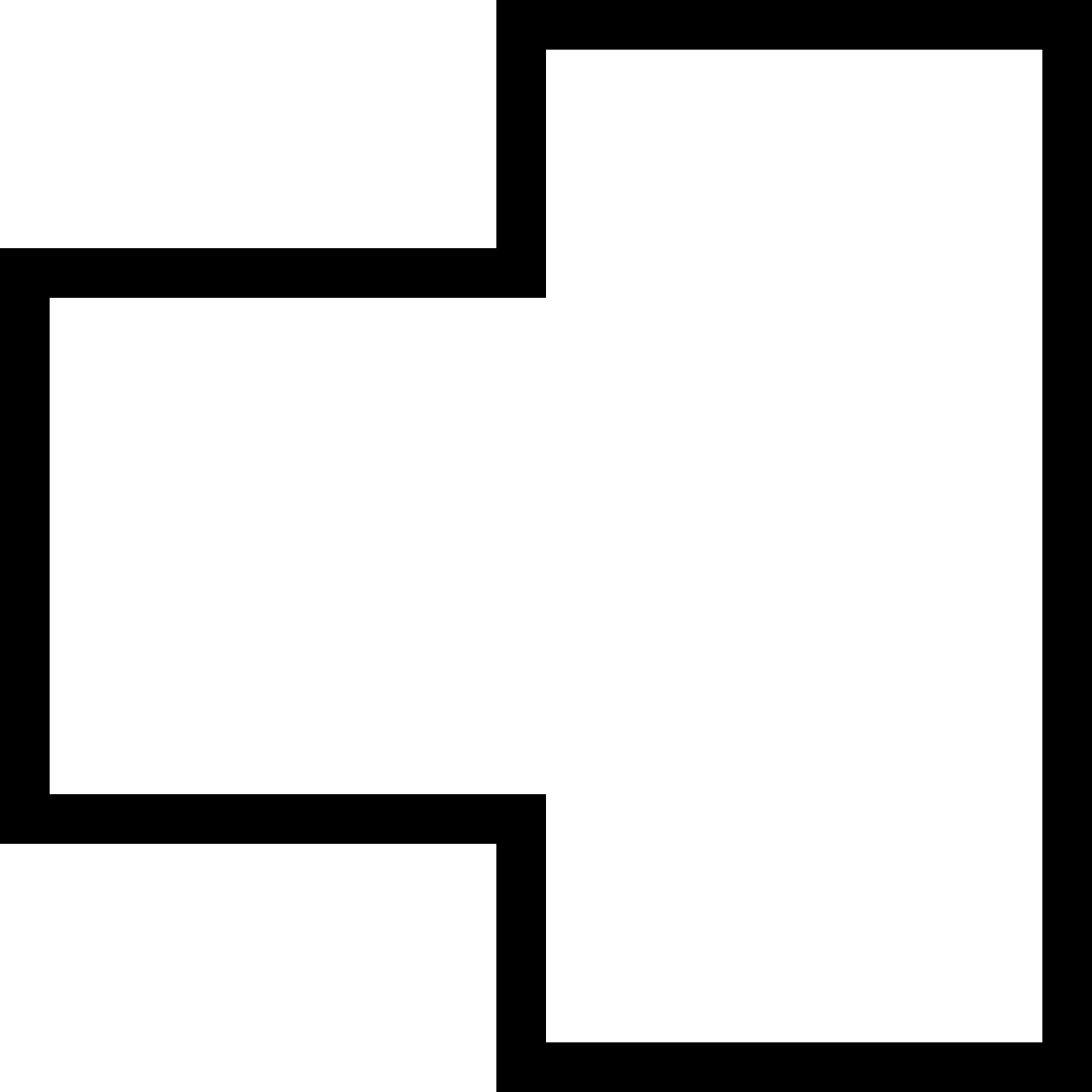} 12~m$^2$ & \includegraphics[width=0.9cm]{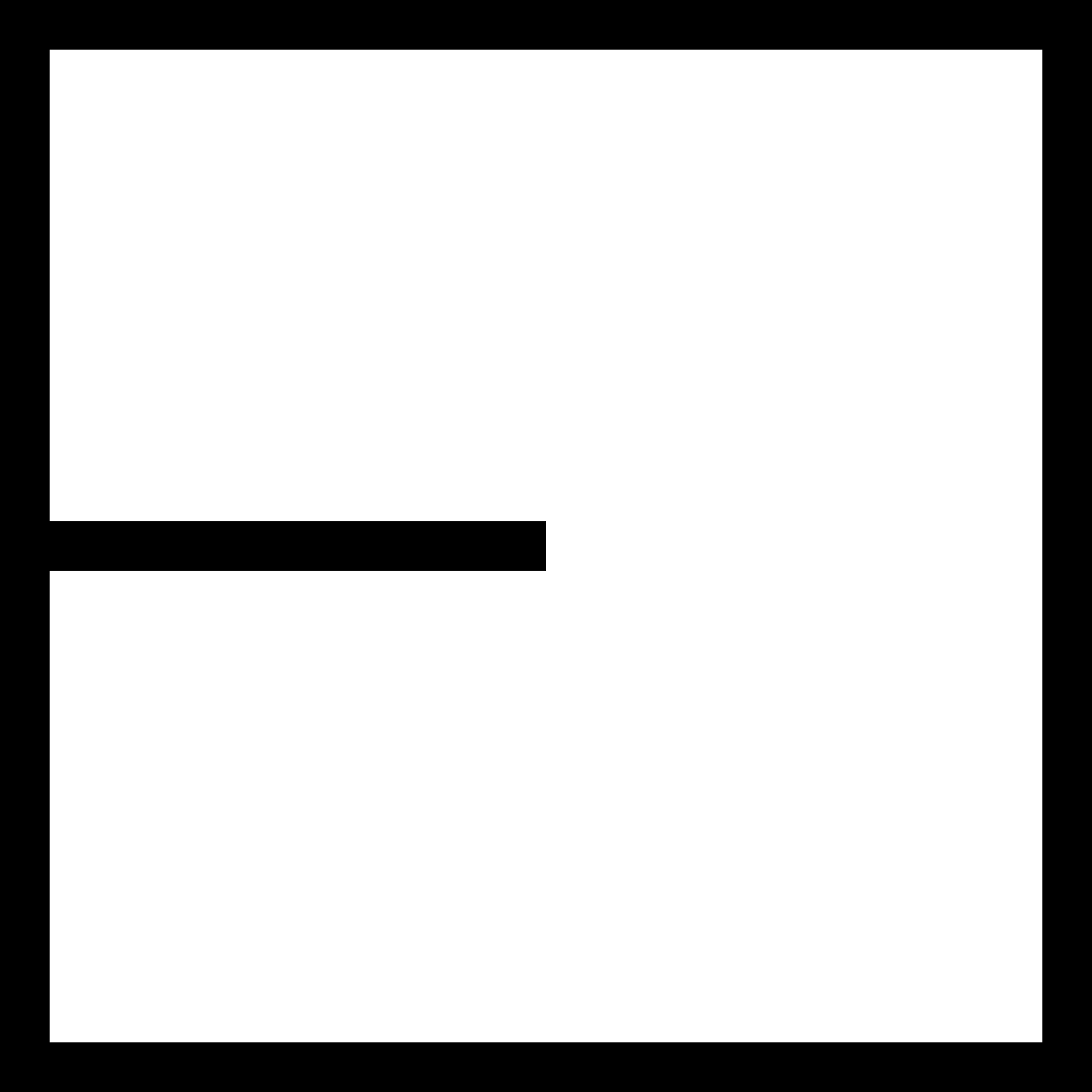} 16~m$^2$ & \includegraphics[width=0.9cm]{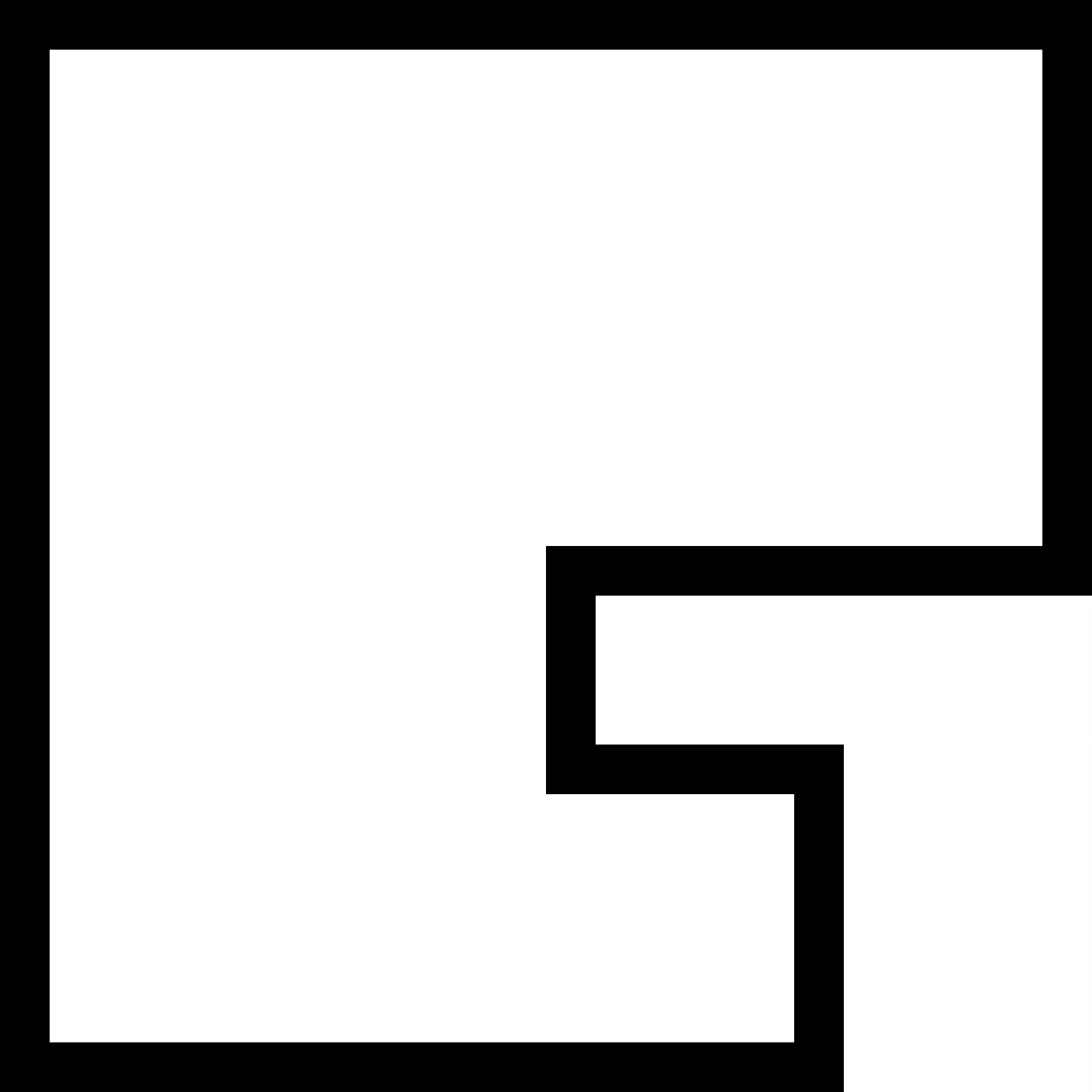} 13~m$^2$ 
\\
\ChangeRT{1pt}
Simulated cover time & 166 $\pm$ 47 & 125 $\pm$ 59 & 265 $\pm$ 131 & 171 $\pm$ 63
\\ 
\hline
Hardware cover time & 76 $\pm$ 22 & 82 $\pm$ 36 & 94 $\pm$ 22 & 128 $\pm$ 42
\\
\hline
KS time for source & 39 $\pm$ 15 \mbox{(52 $\pm$ 16)} & 50 $\pm$ 17 \mbox{(56 $\pm$ 20)} & 60 $\pm$ 37 \mbox{(72 $\pm$ 35)} & 34 $\pm$ 6 \mbox{(40 $\pm$ 0)$^\dagger$}
\\
\end{tabular}
\end{center}
\vspace{-1mm}
\caption{\textbf{Key metrics for laboratory trials.} For the four environments shown, the number of steps to reach full coverage for the source-free case and the number of steps to the first instance of $\log_{10}(P) = -4$ for the source case is reported; each table entry includes the average and standard deviation over 5 trials. For our chosen test parameters, Alg.~\ref{alg:random} only performs a KS test after every 20 steps; the corresponding time to the algorithm terminating is in parentheses. For reference, the experimental configurations were reproduced in PyBullet; the coverage time for 10 simulated trials is reported. $^\dagger$For all 5 source-presence trials in the 13~m$^2$ environment, the KS test triggered after 40 measurements (i.e., the second KS test).}
\label{tab:demo-hardware}
\vspace{-2mm}
\end{table}

As observed in both Figure~\ref{fig:demo-hardware} and Table~\ref{tab:demo-hardware}, if a source was present, Alg.~\ref{alg:random} terminated in fewer steps than necessary to achieve full coverage of the search environment. This was an interesting empirical result which was not theoretically guaranteed, reinforcing that complete coverage is not necessary in order to identify the presence of an anomalous source; complete coverage remains a necessary condition, though, for confirming the absence of sources with high-confidence.

To briefly demonstrate the impact of the detector range, while leveraging the sensing-agnostic nature of Alg.~\ref{alg:random}, we compare the performance of the NaI-equipped robotic inspector to an alternative sensing modality with reduced $r_D$. Three low-efficiency ($r_D = 30~cm$) Geiger counters were evenly distributed around the circumference of the Create 3 platform; we utilized the LND 7314 2in Pancake Geiger detector that is used by the Safecast bGeigie Nano \cite{safecast}. An Adafruit ESP32 Feather V2 was used to count the pulses from each detector individually, and relay the counts to the controller via Bluetooth. The conservative obstacle avoidance of the Create 3, which maintained a buffer of 10~cm around the robot, compounds with the limited range of the Geiger counters, thereby making it very difficult for the robot to remain in the vicinity of an anomalous source for long enough to reach the KS test threshold. 

In a relatively small rectangular space measuring 2$\times$4~m with the gamma sources present, the Geiger counters required 3.33-times more steps than the NaI detector for Alg.~\ref{alg:random} to correctly terminate. Similarly, coverage in the same small space with no source present was achieved 2.54-times faster using the NaI detector than with the Geiger counters, since larger bins (coarser discretization) can be used with the NaI detector. As expected, it is considerably more effective to use a higher efficiency detector.

\section{DISCUSSION AND FUTURE WORK}

We present an algorithm for completing a verification task without requiring, nor revealing, sensitive information from the search environment, defined to include the site layout and any observable features. We derive theoretical guarantees on privacy and correctness that are empirically validated in simulation and in physical hardware experiments.

This work also raises some interesting questions for future exploration. In particular, by more fully characterizing the fundamental trade-off between privacy and time-efficiency for the chosen task, valuable lines of inquiry emerge. Such analysis could uncover optimality conditions for varying privacy regimes. We note that while our procedure admits privacy analysis, it does not, at present, establish optimality. It may also be possible to show that certain privacy and efficiency constraints are mutually incompatible; such ``impossibility results'' provide interesting insights, yet are quite rare in robotics. Exploring the broader spectrum of privacy and efficiency may inform the development of new robotic inspection approaches with varying information constraints.

\section{APPENDIX}

\subsection{Discretization Procedure}
\label{app:discretize}

Consider map $M$ with occupancy function $f'(x, y): [0, l_x] \times [0, l_y] \mapsto \{0, 1\}$ associating all points within the outer bounds to ``free space'' ($0$) or ``not free space'' ($1$). Let $\epsilon_M \geq 0$ be the ``fundamental discretization length'' of map $M$, defined to be the largest $\epsilon \in \mathbb{R}^{\geq 0}$ such that there is a discretization of the occupancy function into bins $\{b_i\}_{i=1}^{N(\epsilon)}$ of side length at least $\epsilon$ with the following properties: (1) the modified, \emph{conservative} occupancy function is $f(i) = \max_{x, y \in b_i} \{f'(x, y)\}$, and (2) the resulting discretized map is traversable by inspector $\mathcal{I}$. Implicitly, this restricts us to maps with $\epsilon_M = \Omega(r_I)$ and a maximum number of bins $N(\epsilon_M) = \mathcal{O}\left(\frac{l_x l_y}{r_I^2}\right)$. The discretization length is also limited by the detector range: $\epsilon_M \lesssim \frac{r_D}{\sqrt{2}}$, where $r_D$ is the distance from which a source is readily detectable, with the corresponding minimum number of bins $N(\epsilon_M) = \Omega\left(\frac{l_x l_y}{r_D^2}\right)$.

\subsection{Mutual Information}
\label{app:mutual}

\begin{proof}
Consider normalized step constants $c_U' = 1$, $c_L' = c_L/c_U$, and similarly normalized step sizes $V_e' = V_e/c_U$ which constitute the realized action distribution. Let $\delta(z) \in (0, 1]$ denote the one-sided significance of the quantile $c(z) = B + z\sqrt{B}$ for a Poisson distribution with mean and variance $B$. That is, for random variables $X_i\overset{\text{iid}}{\sim}$ Poisson$(B)$, $\delta(z) := \mathcal{P}(X >= c(z))$. 

It can be shown that in source-free environments, the step-size density function $q_{\delta(z)}(s \rvert m^-, b)$ is equal to $(c_L' + \delta(z) - c_L'\delta(z))/c_L'$ for $s \in [0, c_L']$ and equal to $(1-\delta(z))$ for $s \in [c_L', 1]$, which is equal to the marginal no-source density $p_{\delta(z)}(s)$. This is due to the uniformity across bins of the count rate distribution. From this, the proof follows by direct decomposition of the mutual information and reduction of the joint probability distribution. By definition,
\begin{equation*}
    \mathcal{MI}(dx_t, M^-) := KL(q_{\delta(z)}(s, M^-) \rVert p_{\delta(z)}(s) p(M^-)).
\end{equation*}

For any $\{s, M^-\}$, $q_{\delta(z)}(s, M^-) = q_{\delta(z)}(s \rvert M^-) p(M^-)$. By the law of iterated expectation, this can be expanded to individual bins in the map as follows: 
\begin{equation*}
    q_{\delta(z)}(s \rvert M^-) p(M^-) = \mathbb{E}_{b \sim \mu_{(M^-)}}\big[ q_{\delta(z)}(s \rvert M^-, b)\big].
\end{equation*}

The latter term is precisely the density function described above, which is independent of $\{M^-, b\}$ and depends solely on $\delta(z)$ (when there is no source). Consequently:
\begin{equation*}
\begin{split}
    q_{\delta(z)}(s \rvert M^-) & = \mathbb{E}_{b \sim \mu_{(M^-)}}\big[ q_{\delta(z)}(s \rvert M^-, b)\big] \\
    & = q_{\delta(z)}(s \rvert M^-, b) \\
    & = p_{\delta(z)}(s).
\end{split}
\end{equation*}

Therefore, $q_{\delta(z)}(s \rvert M^-) = p_{\delta(z)}(s)$, and we recover the expression for the mutual information, which yields the desired result: 
\begin{equation*}
    \begin{split}
        \mathcal{MI}(dx_t, M^-) & := KL(q_{\delta(z)}(s, M^-) \rVert p_{\delta(z)}(s) p(M^-)) \\
        & = KL(q_{\delta(z)}(s \rvert M^-)p(M^-) \rVert p_{\delta(z)}(s) p(M^-)) \\
        & = KL(p_{\delta(z)}(s) p(M^-) \rVert p_{\delta(z)}(s) p(M^-)) \\
        & = 0.
    \end{split}
\end{equation*}
\end{proof}

Importantly, the resulting zero mutual information is true for \emph{any} probability law over the set of source-free maps. In particular, this holds uniformly across all maps, which can be seen by taking the set of probability laws that place all of their mass on a particular map.

\subsection{Exponential Family}
\label{app:exponential}

To the best of our knowledge, this result is not a trivial application of eigenvalue theory for directed graphs, nor of Perron-Frobenius theory on graph transition matrices. We believe that the conductance arguments of \parencite{mihail_conductance_1989} apply here to give tighter bounds by using the local structure of the physically grounded random walk. For completeness, in case the above results do not apply in an edge case, we assume a looser lower bound. 

\begin{proof}
By assumption, we are tasked with exploring a contiguous (i.e., traversable) space; this translates to the resulting graph representation being strongly connected. Furthermore, because our algorithm allows for small step sizes (down to zero), the probability of self-loops is positive (non-zero) for every node. As such, the graph representation is necessarily aperiodic. From this, first, Perron-Frobenius theory gives that for a strongly connected, aperiodic graph, the associated transition matrix has a unique, positive stationary distribution. Denote this as $\pi_G > 0$. Second, from strong connectivity, the graph diameter ($D_G$) is less than or equal to $N$. It follows that, for any node $i$, the radius of the graph $r_G(i)$ (the maximal distance of any node $j \neq i$ to node $i$) is necessarily greater than zero and less than or equal to $D_G$. 

The remainder of the proof is relatively straightforward, as follows. Set the output of a state $i$ (the i$^{th}$ column of $\mathbb{P}$, using the convention $\pi_G = \mathbb{P}\pi_G$) to be uniformly zero, such that $i$ is, essentially, an absorbing node. Let the path from the node farthest from $i$ (denote this as $j^*(i)$) to $i$ be defined as $\{j_k\}_{k=0}^{r_G(i)}$, where $j_0 = j^*(i)$ and $j_{r_G(i)} = i$. Then after $r_G(i)+1$ time steps, for any $v_0 \in \Delta_N$, $\lVert \mathbb{P}^{r_G(i)+1}v \rVert_1 \leq 1-c(i)$, where $c(i) = \Pi_{k=1}^{r_G(i)} \mathbb{P}_{j_k \leftarrow j_{k-1}}$. Let $\lambda(i) = 1 - c(i) < 1$, and denote a non-dimensionalized time $\tau = t/r_G(i)$. We can interpret the one-norm of the vector $v_\tau(v_0)$ as the upper bound of the probability of not having reached state $i$ in time $\tau$, given that we have started in (proper, normalized) state distribution $v_0$. This function has the form $F_0(\tau) = \lambda^\tau$. It is, therefore, clear that the (lower bound) cumulative distribution function of the first passage time is $F(t) = 1-F_0(\tau) = 1-\lambda(i)^{\tau} = 1-\lambda(i)^{t/r}$. Because $t$ is discrete, this can be modeled as a geometric distribution in non-dimensionalized time $\tau$, or in $\lfloor \frac{t}{r_G(i)} \rfloor$. 
\end{proof}

\subsection{Improved High-Probability Bounds on Coverage Time}
\label{app:convolution}

We describe a means of generating high-probability bounds on coverage time in the detection application, demonstrating (consistent with the literature) that coverage is dominated by the worst-case cover time of a single node. Furthermore, we show that significantly better bounds (empirically) can be obtained by leveraging the exponential family properties of the cover time distribution.

In essence, the method utilizes an informative prior: reaching adjacent nodes should have relatively strong correlation. Consider a physical, discretized map with $N$ nodes in a set $v_n$. Partition the nodes into $N/2$ disjoint subsets $b_i = \{v_{n_i, 1}, v_{n_i, 2}\}$ of two nodes each (in this and all ensuing steps, if the number of input nodes is odd, there will be $(N+1)/2$ output subsets, where the last subset is singleton and will be incorporated in a subsequent step). Define for each output subset, $b_i$, the following conditional event: the output subset $b_i$ is fully explored at or before time $t_0 + \tau$ with probability greater than $1-\delta/(N)$, given that at least one node within $b_i$ has been reached at time $t_0$. This is equivalent, for each subset, to $\max\{\bar{t}_{1 \rightarrow 2}, \bar{t}_{2 \rightarrow 1}\}$, the respective times to traverse from the first node in $b_i$ to the second, and from the second to the first, each with probability $1-\delta/(2N)$. Associate the above maximum time with bin $b_i$; by a union bound over the two possibilities, the maximum time holds with probability $1-\delta/N$. Note that there are $N/2$ such bins. 

Now, repeat the process using the previous output subset as the input nodes with a minor adjustment to account for accumulated traversal time. That is, bins $b_i$ are the $N/2$ input ``nodes,'' and we can let $b_j$ be the $N/4$ output subsets. Let $b_j = \{b_1^{[j]}, b_2^{[j]}\}$; now, the combination requires us to consider the maximum over four possibilities: (1) exploring $b_1^{[j]}$, (2) exploring $b_2^{[j]}$, (3) going from $b_1^{[j]}$ to $b_2^{[j]}$ and then exploring $b_2^{[j]}$, and (4) going from $b_2^{[j]}$ to $b_1^{[j]}$ and then exploring $b_1^{[j]}$. The first two exploration possibilities are strictly dominated by the last two, so with each $b_j$ we associate the value $\max\{\bar{t}_{1 \rightarrow 2} + \bar{t}_2, \bar{t}_{2 \rightarrow 1} + \bar{t}_1 \}$. This is the same as the first step, except that in the first step the associated ``accumulated times'' (i.e., the time traveling from one node to the next) were zero, and left implicit. Assuming that the traversals are bounded again by probability $1-\delta/(2N)$, the union bound (combined with the existing ones from preceding steps) bounds the outcomes for the further-reduced $b_j$.

Inductively, repeating this process for $\lceil \log_2(N) \rceil$ steps is guaranteed to result in a single remaining bin corresponding to the entire contiguous free space. Summing over combination steps, we have that there are $N$ combinations made ($N/2 + N/4 + ...$) and that each combination adds two union bounds. Therefore, there are $2N$ union bounds, and the resulting, consolidated ``maximum of maxima'' is an upper bound on the coverage time with probability greater than $1-\delta$. As such, the resulting high-confidence upper bound is $\mathcal{O}(\bar{T}\log{N})$, where $\bar{T}$ is the worst case (over the $N$ ``real'' nodes) upper bounded cover time for a single node. 

A key intuition that we rely upon is the actual realization of the bin partition; physical proximity is a strong prior for the mixing of physical random walks, which allows for useful speed-up. Concretely, we can construct a partition of pairs of physically adjacent bins which mixes faster than a partition of randomly selected pairs of bins. Furthermore, this intuition also suggests that an optimal step size (from the perspective of obtaining upper bounds) should exist and that it should not be too large, as a large step size -- beyond taking more time to execute -- weakens the prior over good partition pairs. Therefore, this reflects the implicit trade-off between quick exploration at large scales (which benefits from larger step sizes) and thorough exploration at smaller scales (which benefits from smaller step sizes).

\section*{ACKNOWLEDGMENTS}

This work has been supported by the National Science Foundation Graduate Research Fellowship Program under Grant No. DGE-2039656, and the Consortium for Monitoring, Technology, and Verification under Department of Energy National Nuclear Security Administration award number DE-NA0003920.

\section*{REFERENCES}
\widowpenalty=-1000
\clubpenalty=-1000
\looseness=-1000
\printbibliography[heading=none]

\end{document}